\documentclass{article}
\usepackage{amsmath,amssymb}
\usepackage{graphicx}
\usepackage{enumitem}
\usepackage{booktabs}
\usepackage{multirow}
\usepackage{algorithm}
\usepackage{algpseudocode}
\usepackage{caption}
\usepackage{subcaption}

\PassOptionsToPackage{sort&compress}{natbib}
\usepackage[preprint]{anonymous_2026} 

\newif\ifincludeappendix
\includeappendixtrue
\usepackage{xcolor}

\title{Object-Centric Residual RL for Zero-Shot Sim-to-Real VLA Enhancement}

\author{
  Kinam Kim$^{1,2,\dagger}$, Namiko Saito$^{2}$, Heecheol Kim$^{2}$, Katsushi Ikeuchi$^{2,3}$, Jaegul Choo$^{1}$, Yasuyuki Matsushita$^{2}$\\
  $^{1}$ KAIST, South Korea,
  $^{2}$ Microsoft Research Asia - Tokyo, Japan \\
  $^{3}$ The University of Tokyo, Japan \\
}

\begin{document}

\maketitle
\renewcommand{\thefootnote}{\fnsymbol{footnote}}
\footnotetext[2]{Work done during an internship at Microsoft Research Asia.}
\renewcommand{\thefootnote}{\arabic{footnote}}
\begin{center}
    \includegraphics[width=\textwidth]{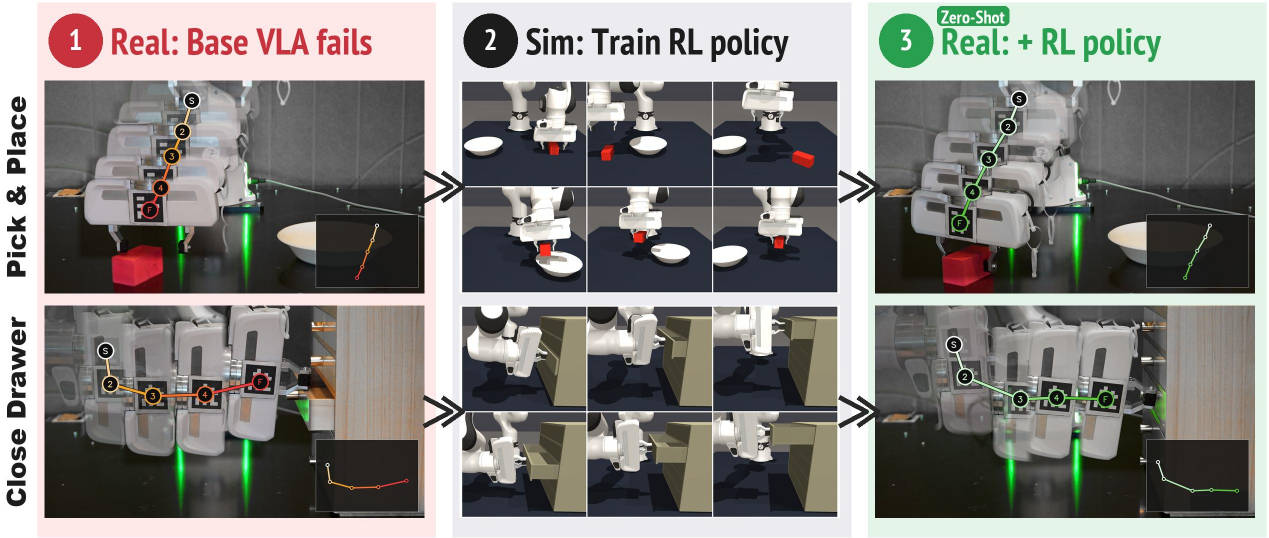}
    \captionof{figure}{\textbf{Object-centric residual RL for zero-shot sim-to-real VLA enhancement.}
    The base VLA fails on the real robot (\textbf{\textcolor{red}{left}}). A residual policy trained purely in simulation \textbf{(middle)} is added zero-shot to recover task success on the same real-robot setup (\textbf{\textcolor{green!50!black}{right}}).
    }
    \label{fig:teaser}
\end{center}


\begin{abstract}
    Vision-Language-Action (VLA) models can generalize across diverse manipulation tasks, but their imitation-learning-based policies remain brittle in precise physical interactions due to compounding execution errors; \textit{Can a reinforcement learning policy trained purely in simulation improve the robustness of real-world VLAs zero-shot?} Residual RL, which learns a corrective policy on top of a frozen VLA, offers a natural framework, but existing approaches face a fundamental sim-to-real dilemma: privileged-state methods require lossy distillation for deployment; image-based methods suffer from the visual domain gap; and real-world RL is costly and unsafe. We propose an object-centric residual RL framework that refines VLA actions using object poses, enabling a compact observation space that transfers consistently between simulation and reality.
    To align the two domains, we additionally replay the same teleoperation demonstrations in simulation to train a sim counterpart of the real-world VLA. The residual RL policy is trained only in simulation with pose noise injection and dropout, and transfers zero-shot to the real robot.
    Across five manipulation tasks on a real Franka Research 3 (FR3) robot, our method improves the success rate from $42\%$ to $\mathbf{76\%}$ zero-shot, and the improved rollouts can be further reused to retrain the base VLA for self-improvement without additional teleoperation.
    Project page: \url{https://www.microsoft.com/en-us/research/articles/object-centric-residual-rl/}
\end{abstract}



\section{Introduction}
\label{sec:introduction}

Vision-Language-Action models (VLAs) enable broad manipulation capabilities by leveraging large-scale pretraining and robot demonstrations~\cite{brohan2023rt1, brohan2023rt2, openx2024, octo2024, kim2024openvla, black2024pi0, pi05_2025, bjorck2025gr00t}.
However, because VLAs are trained via imitation learning, small errors accumulate over time and lead to failures in unseen states~\cite{ross2011dagger, zhao2023act}.
Reinforcement learning (RL) can improve recovery through online interaction, but directly applying RL to modern VLAs is difficult because many architectures rely on diffusion~\cite{ho2020ddpm, chi2023diffusionpolicy} or flow matching~\cite{lipman2023flow, black2024pi0} for action generation, whose iterative denoising is not readily differentiable through standard policy gradients.
Residual RL~\cite{silver2018residual, johannink2019residual} addresses this by learning a lightweight corrective policy on top of a frozen VLA, but sim-to-real transfer of the residual remains a major challenge.

Three approaches have been pursued regarding residual RL, each with a distinct failure mode.
\textbf{Distillation-based residual RL}~\cite{resip2024} trains on privileged simulator state and requires teacher-student \emph{distillation} into an image-based student for deployment, incurring performance loss.
\textbf{Image-based sim residual RL} avoids distillation by operating directly on images, but suffers from a large visual sim-to-real domain gap that prevents zero-shot transfer of the sim-trained residual.
\textbf{Real-world residual RL}~\cite{resfit2025, pld2025} eliminates the need for sim-to-real transfer by training directly on the real robot, but is costly and raises safety concerns~\cite{dulac2021challenges}.
None of these paradigms achieves zero-shot transfer of a sim-trained residual policy to a real robot.

In this work, we observe that prior sim-trained residual policies fail to transfer because their observation spaces are inherently domain-dependent: privileged-state methods~\cite{resip2024} rely on quantities unavailable on the real robot and must distill into an image policy, while image-based methods face a large visual sim-to-real gap.
We take a different approach: rather than explicitly bridging them, we substantially reduce the discrepancies seen by the residual policy
by building the residual on observations that are consistently recoverable in both domains---6-DoF object poses, proprioceptive state, and the base VLA action.
Object pose can be reliably obtained via off-the-shelf estimators~\cite{wen2024foundationpose, ravi2024sam2}, and because the residual operates on this low-dimensional state rather than images, it transfers \textbf{zero-shot} without distillation or real-world RL.
To align action distributions across domains, we replay the same teleoperation trajectories in simulation to train a sim VLA alongside the real one.

Beyond zero-shot deployment, our framework also enables automatic VLA self-improvement. Successful real-robot rollouts collected by deploying the residual-corrected policy can be aggregated across tasks to retrain a single multi-task VLA, producing higher-quality training data without any additional teleoperation.

Our contributions are as follows:
\begin{itemize}[nosep]
    \item We propose a \textbf{zero-shot sim-to-real residual RL framework} with two design choices: paired sim and real VLAs aligned via teleoperation replay, and a \textbf{domain-invariant residual interface} that sidesteps the sim-to-real gaps without distillation or real-world RL.
    \item We show that real-robot rollouts from the residual-corrected policy can be aggregated across tasks to retrain a single \textbf{multi-task VLA}, enabling \textbf{automatic self-improvement} without additional teleoperation.
    \item We demonstrate consistent improvement across \textbf{five manipulation tasks} in both simulation and \textbf{real-world zero-shot deployment}, with an average improvement from $42\%$ to $\mathbf{76\%}$ success rate on a real FR3 robot.
\end{itemize}

\section{Related Work}
\label{sec:related_work}

\paragraph{Residual Reinforcement Learning.}
Residual RL~\cite{silver2018residual, johannink2019residual} trains a corrective policy on top of a frozen base, combining the generalization of the base with the precision of RL.
ResiP~\cite{resip2024} trains the residual policy on privileged simulator state, achieving strong sim performance but requiring a separate teacher-student distillation step for real-world deployment, which incurs a non-trivial performance loss during the transfer. 
RialTo~\cite{rialto2024} avoids real-world RL by constructing a digital twin via 3D scanning of the entire workspace and training point-cloud-based policies through an inverse distillation pipeline, but the end-to-end process (3D reconstruction, scene registration, point-cloud policy training, and inverse distillation) incurs substantial per-task time and engineering overhead, limiting scalability to new tasks.
ResFiT~\cite{resfit2025} operates directly on images and proprioception, but must train on the real robot, requiring safe exploration infrastructure, episode resets, and real-world reward detection.
PLD~\cite{pld2025} extends this to VLAs with a three-stage probe-learn-distill pipeline, also requiring real-world RL and an additional reward classifier.
RPD~\cite{rpd2025} distills VLA knowledge into an RL student entirely in simulation, but has not demonstrated real-world deployment.
In contrast, our method operates on object pose, a compact representation accurately recoverable in reality, enabling zero-shot sim-to-real transfer without distillation, real-world RL, or complex reconstruction pipelines.


\paragraph{Sim-to-Real Transfer.}
Bridging the gap between simulation and reality has been a longstanding challenge~\cite{zhao2020simtoreal_survey}.
Domain randomization~\cite{tobin2017domain, peng2018simtoreal, handa2023dextreme} varies visual and physical parameters during training to promote robustness, and has enabled impressive sim-to-real results including dexterous in-hand manipulation~\cite{andrychowicz2020dexterous}.
Recent work automates this process with language models~\cite{ma2024dreureka}, while others adapt randomization distributions using real-world data~\cite{chebotar2019closing, ramos2019bayessim} or learn residual corrections from online human feedback~\cite{jiang2024transic}.
When the policy uses privileged simulator state, a common recipe is teacher-student distillation~\cite{hinton2015distilling, resip2024}, where a state-based teacher is first trained in simulation and then distilled into a vision-based student for real-world deployment.
We take an orthogonal approach: instead of aligning visual or dynamics distributions across domains, we design the residual's observation space to remain consistent between simulation and reality, enabling zero-shot transfer without domain adaptation or distillation.

\section{Method}
\label{sec:method}

\begin{figure*}[t]
    \centering
    \includegraphics[width=\textwidth]{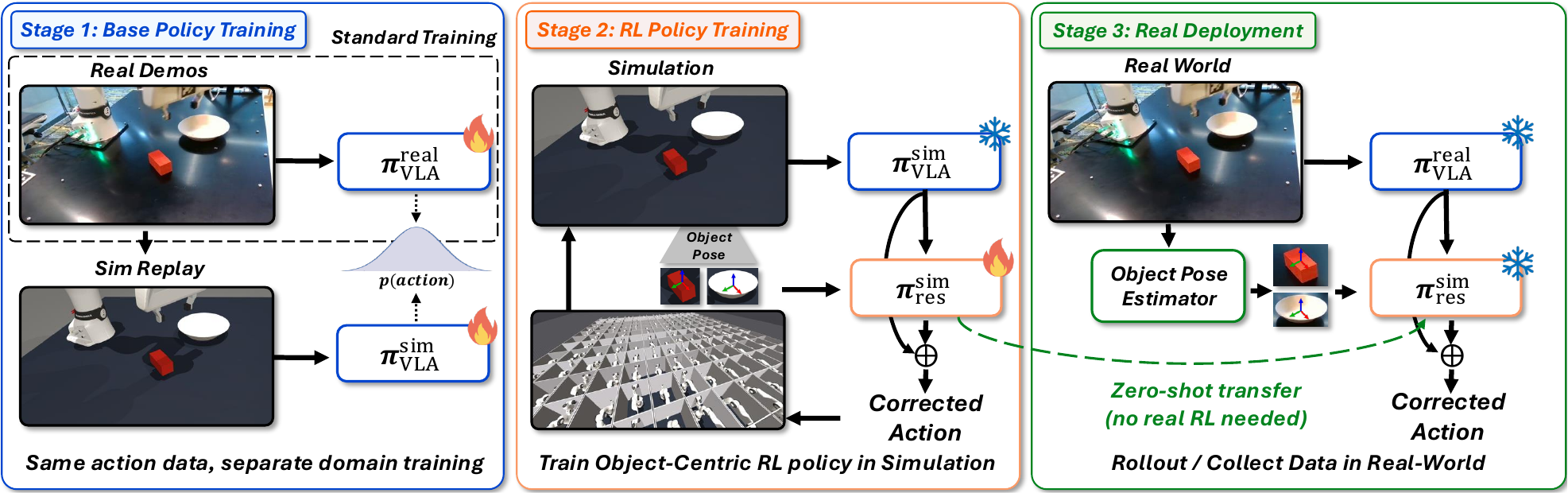}
    \caption{Overview of the object-centric residual RL pipeline.}
    \label{fig:pipeline}
\end{figure*}

Given a VLA trained on real-robot demonstrations,
our framework enhances it with a sim-trained residual policy through three stages (Fig.~\ref{fig:pipeline}): (1) building a simulation counterpart of the VLA via teleoperation replay (Section~\ref{sec:stage1}), (2) training an object-centric residual policy in simulation (Section~\ref{sec:stage2}), and (3) zero-shot deployment on the real robot (Section~\ref{sec:deployment}). The combined deployment action is
\begin{equation}
    a_t = a_t^{\text{base}} \;\oplus\; \pi_{\text{res}}^{\text{sim}}(s_t),
    \label{eq:residual}
\end{equation}
where $a_t^{\text{base}}$ is the current base action drawn from the real VLA's chunked rollout (Stage 1), and $\pi_{\text{res}}^{\text{sim}}$ is the sim-trained residual queried at every timestep $t$ on object-centric observation $s_t$ (Stage 2).
The VLA takes an RGB observation, proprioceptive state, and language instruction; the VLA is kept frozen during residual training, and both are frozen at deployment.
$\oplus$ denotes per-component action composition: addition for position and gripper, quaternion multiplication for rotation.
The residual is trained to consistently improve task success over $\pi_\mathrm{VLA}$ alone, and deploys zero-shot on the real robot without any real-world RL or distillation.
Beyond deployment, the residual-corrected policy can also be used to retrain the base VLA itself, enabling multi-task generalization of per-task residual-RL behaviors (Section~\ref{sec:self-improvement}).

\subsection{Stage 1: Paired Sim/Real VLA via Teleoperation Replay}
\label{sec:stage1}

Standard task-specific VLA fine-tuning collects teleoperation demonstrations on the real robot and trains a VLA on real (RGB, action) pairs. We extend this standard recipe by additionally replaying the same teleoperation actions in a simulation environment~\cite{mandlekar2024mimicgen}, rendering a parallel sim dataset that trains a sim VLA $\pi_\mathrm{VLA}^{\mathrm{sim}}$ alongside the standard real VLA $\pi_\mathrm{VLA}^{\mathrm{real}}$. Because both VLAs are supervised by identical teleoperation actions, they learn aligned action distributions despite seeing different visual domains, letting the residual train alongside $\pi_\mathrm{VLA}^{\mathrm{sim}}$ in simulation and transfer to $\pi_\mathrm{VLA}^{\mathrm{real}}$ zero-shot at deployment.
In short, $\pi_\mathrm{VLA}^{\mathrm{sim}}$ provides $a_{\text{base}}$ for residual RL training in simulation, while $\pi_\mathrm{VLA}^{\mathrm{real}}$ replaces it at deployment.
Since the residual policy does not observe images, the simulation need not be visually realistic, significantly reducing the engineering effort required to construct the simulation environment.


\subsection{Stage 2: Object-Centric Residual RL}
\label{sec:stage2}

The key challenge is choosing the observation $s$ for $\pi_{\text{res}}^{\text{sim}}$.
If $s$ contains privileged simulator state (e.g., contact forces), transfer fails.
If $s$ contains images, the visual domain gap prevents zero-shot transfer.
We resolve this by constructing $s$ from quantities that are both informative and accurately recoverable in reality.


\paragraph{Observation space.}
\label{sec:observation}
In our work, the residual policy observation is defined as:
\begin{equation}
    s_t = [\,s_t^{\text{obj}},\; s_t^{\text{prop}},\; a_t^{\text{base}}\,],
    \label{eq:obs}
\end{equation}
where $s_t^{\text{obj}}$ is the 6-DoF pose (position and orientation) of the task-relevant objects (manually specified in the current setup), $s_t^{\text{prop}}$ is the proprioceptive state, and $a_t^{\text{base}}$ is the current base action drawn from the VLA's action chunk.
For tasks involving multiple objects (e.g., Stack Cube requires tracking both the grasped cube and the target cube), the poses of all task-relevant objects are concatenated into $s_{\text{obj}}$.
Our residual formulation enables continued improvement via RL while the VLA remains frozen, decoupling the pose-based correction from the base policy's training.
We note that our approach is orthogonal to recent work on 3D-conditioned VLAs~\cite{ze2024dp3, ke2024diffuseractor, qu2025spatialvla}: if such a model serves as the base policy, our residual module can still be applied on top to provide additional RL-based corrections.
Note that the framework is also agnostic to the specific proprioceptive representation (e.g., end-effector pose vs.\ joint angles) and to the action parameterization of the base VLA.

\paragraph{Zero-shot transfer condition.}
\label{sec:zero-shot}
We formalize the condition under which a sim-trained residual policy can transfer zero-shot.
Let $s^{\text{sim}}$ and $s^{\text{real}}$ denote the observation vectors in simulation and reality, respectively.
Zero-shot transfer becomes feasible when the residual policy is robust to the deployment noise $\mathcal{P}_\eta$:
\begin{equation}
    s_t^{\text{real}} = s_t^{\text{sim}} + \eta_t, \quad \eta_t \sim \mathcal{P}_\eta.
    \label{eq:zero-shot}
\end{equation}
The magnitude of $\mathcal{P}_\eta$ depends on the choice of observation $s_t$.
If $s_t$ includes privileged simulator state (e.g., contact forces), $\eta_t$ is large due to systematic physics modeling errors.
If $s_t$ includes images, $\eta_t$ is large due to rendering discrepancies in lighting, textures, and backgrounds.
Our observation space minimizes $\mathcal{P}_\eta$ by construction:
\begin{itemize}[nosep]
    \item Proprioception (end-effector pose and gripper state): domain-invariant, contributing $\eta \approx 0$.
    \item Base VLA action: $\pi_\mathrm{VLA}^{\mathrm{sim}}$ and $\pi_\mathrm{VLA}^{\mathrm{real}}$ are trained on the same teleoperated trajectories, yielding closely aligned outputs ($\eta \approx 0$).
    \item Object pose: estimated via FoundationPose~\cite{wen2024foundationpose} with SAM2~\cite{ravi2024sam2}; the only component with non-negligible $\eta_t$.
\end{itemize}
We address the remaining pose estimation noise $\mathcal{P}_\eta$ by augmenting the pose input with structured noise during training, as detailed below.

\paragraph{Robust object pose training.}
\label{sec:robustness}
Real-world pose estimators introduce noise and occasional failures.
To ensure that the residual policy is robust to these errors, we apply two forms of augmentation during training.
We decompose the 6-DoF object pose $s_{\text{obj}}$ into position $p_{\text{obj}}$ and orientation $q_{\text{obj}}$ components.
First, at each timestep we perturb each component with random noise:
\begin{equation}
    \hat{p}_{\text{obj}} = p_{\text{obj}} + \epsilon_p, \quad \hat{q}_{\text{obj}} = q_{\text{obj}} \otimes \epsilon_q,
    \label{eq:noise}
\end{equation}
where $\otimes$ denotes quaternion multiplication, each component of $\epsilon_p$ is independently sampled from $\mathcal{U}(-\tilde{\sigma}_p, \tilde{\sigma}_p)$ with $\tilde{\sigma}_p \sim \mathcal{U}(0, \sigma_p^{\max})$ resampled per timestep, and $\epsilon_q$ is a small random rotation with magnitude similarly drawn via $\tilde{\sigma}_q \sim \mathcal{U}(0, \sigma_q^{\max})$.
This hierarchical uniform sampling exposes the policy to a continuous range of small pose perturbations, mirroring the varying accuracy of real-world pose estimators.
We denote the combined noise-augmented pose as $\hat{x}_{\text{obj}} = (\hat{p}_{\text{obj}}, \hat{q}_{\text{obj}})$.
Second, with probability $\rho_{\text{drop}}$ we zero out the entire object pose vector:
\begin{equation}
    \tilde{x}_{\text{obj}} =
    \begin{cases}
        \hat{x}_{\text{obj}} & \text{with probability } 1 - \rho_{\text{drop}} \\
        \mathbf{0} & \text{with probability } \rho_{\text{drop}}.
    \end{cases}
    \label{eq:dropout}
\end{equation}
This forces the policy to learn a fallback strategy using only proprioception and the base action, ensuring graceful degradation when the pose estimator fails.

\paragraph{Reinforcement learning.}
\label{sec:training}
We train the residual policy using TD3~\cite{fujimoto2018td3} with clipped exploration noise for stable off-policy optimization.
Exploration adds clipped Gaussian noise on top of the combined action, while pose-noise scales $(\sigma_p^{\max}, \sigma_q^{\max})$ and dropout probability $\rho_{\text{drop}}$ control the augmentation magnitudes.
$\pi_\mathrm{VLA}^{\mathrm{sim}}$ is queried every $H$ steps to produce an $H$-length action chunk
\begin{equation}
    A_k = \pi_\mathrm{VLA}^{\mathrm{sim}}(o_{kH}^{\text{img}}, s_{kH}^{\text{prop}}, l), \quad k = \lfloor t/H \rfloor,
    \label{eq:chunk}
\end{equation}
where $o_{kH}^{\text{img}}$ and $s_{kH}^{\text{prop}}$ are the RGB observation and proprioceptive state at chunk timestep $kH$, and $l$ is the language instruction.
From the resulting chunk $A_k$, we read the current base action $a_t^{\text{base}} = A_k[t \bmod H]$. The residual $\pi_{\text{res}}^{\text{sim}}$ is queried at every timestep $t$ with observation $s_t$ from Eq.~(\ref{eq:obs}), yielding the combined action:
\begin{equation}
    a_t = a_t^{\text{base}} \;\oplus\; \pi_{\text{res}}^{\text{sim}}(s_t).
    \label{eq:train}
\end{equation}
The residual policy is trained with dense rewards (see Appendix~\ref{sec:app-reward} for details).

\subsection{Stage 3: Zero-Shot Deployment}
\label{sec:deployment}

At deployment time, both $\pi_\mathrm{VLA}^{\mathrm{real}}$ and $\pi_{\text{res}}^{\text{sim}}$ are frozen; the system requires no further training or adaptation. The base action $a_t^{\text{base}}$ is read from $\pi_\mathrm{VLA}^{\mathrm{real}}$'s action chunk analogously to Eq.~(\ref{eq:chunk}), and the combined action follows Eq.~(\ref{eq:train}).
At each timestep, we additionally consult the pose estimator's tracking confidence $c_t$: when $c_t$ falls below a threshold $\tau_c$, the pose input is zeroed out to trigger the dropout fallback learned during training (Eq.~(\ref{eq:dropout})).
The residual module adds minimal computational overhead: the actor forward pass takes less than 1\,ms, and FoundationPose runs in tracking mode after initial registration via SAM2~\cite{ravi2024sam2} (${\sim}$18\,ms per frame); we run it asynchronously so that pose estimation does not bottleneck the control loop.
The confidence-gated dropout bridges training and deployment: the random dropout at training time prepares the policy for the \emph{systematic} dropout that occurs at deployment when poses are lost.

\subsection{VLA Self-Improvement}
\label{sec:self-improvement}

Beyond deployment-time enhancement, successful rollouts from the residual-corrected policy can be merged with the original demonstrations to retrain the base VLA via supervised fine-tuning.
This self-improvement loop requires no additional data collection with teleoperation; rollout data from multiple task-specific residuals can be aggregated to retrain a single multi-task VLA, preserving generalist ability.

\section{Experimental Setup}
\label{sec:setup}


\begin{figure*}[t]
    \centering
    \includegraphics[width=\textwidth]{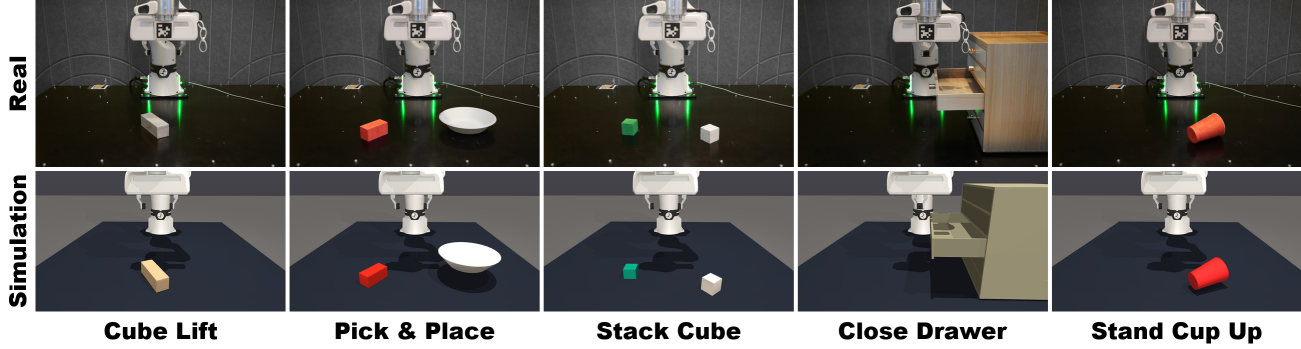}
    \caption{Real (top) and simulated (bottom) environments for all five evaluation tasks.}
    \label{fig:tasks}
\end{figure*}

We evaluate on five tabletop manipulation tasks, instantiated both in MuJoCo~\cite{todorov2012mujoco} simulation (for residual training) and on a real FR3 robot (for zero-shot deployment):

\begin{enumerate}[nosep]
    \item \textbf{Cube Lift (Lift)}: Grasp a cube from the table and lift it $3$\,cm.
    \item \textbf{Pick-and-Place (PnP)}: Pick up a cube and place it into a bowl.
    \item \textbf{Stack Cube (Stack)}: Pick up a white cube and stack it on top of a green cube.
    \item \textbf{Close Drawer (Close)}: Push an open cabinet drawer closed.
    \item \textbf{Stand Cup Up (Stand)}: Grasp a cup lying on its side and stand it upright.
\end{enumerate}

The simulation is built by measuring real object dimensions; visual realism is not required (Fig.~\ref{fig:tasks}). Further details on the simulation environment are provided in Appendix~\ref{sec:app-sim-construction}.
We use GR00T-N1.5~\cite{bjorck2025gr00t}, an open-source VLA, as the base policy fine-tuned on 30 teleoperation demonstrations per task for each domain (sim and real)~\cite{mandlekar2021robomimic}.
The residual policy is a lightweight 2-layer MLP trained with TD3~\cite{fujimoto2018td3}; a single forward pass takes ${\sim}0.06$\,ms on GPU, less than $0.05\%$ of the VLA's ${\sim}140$\,ms inference time.

\section{Results}
\label{sec:result}

We design our experiments to answer the following questions:
\begin{enumerate}[nosep]
    \item Does the residual improve the base VLA zero-shot on a real robot? (Section~\ref{sec:main-results})
    \item What observation and robustness designs enable sim-to-real transfer? (Section~\ref{sec:ablation})
    \item When and how does the residual intervene? (Section~\ref{sec:analysis})
    \item Can residual-corrected rollouts bootstrap VLA self-improvement? (Section~\ref{sec:sft})
\end{enumerate}

\subsection{Main Results}
\label{sec:main-results}

\begin{figure*}[t]
\centering
\begin{minipage}[t]{0.55\textwidth}
  \centering
  \vspace{0pt}
  \captionof{table}{Success rates in simulation and real-robot. Simulation results are reported as mean $\pm$ standard deviation over 3 seeds.}
  \label{tab:main-results}
  \resizebox{\linewidth}{!}{%
  \begin{tabular}{l cc cc}
  \toprule
   & \multicolumn{2}{c}{Simulation} & \multicolumn{2}{c}{Real Robot} \\
  \cmidrule(lr){2-3} \cmidrule(lr){4-5}
  Task & Base & +Res. & Base & +Res. \\
  \midrule
  Cube Lift      & $4.3/20{\scriptstyle\,\pm 0.6}$   & $\mathbf{20.0/20}{\scriptstyle\,\pm 0.0}$   & $7/20$   & $\mathbf{17/20}$   \\
  Pick-and-Place & $7.0/20{\scriptstyle\,\pm 2.6}$   & $\mathbf{17.0/20}{\scriptstyle\,\pm 2.0}$   & $9/20$   & $\mathbf{16/20}$   \\
  Stack Cube     & $10.0/20{\scriptstyle\,\pm 1.0}$  & $\mathbf{14.7/20}{\scriptstyle\,\pm 0.6}$   & $7/20$   & $\mathbf{15/20}$   \\
  Close Drawer   & $11.3/20{\scriptstyle\,\pm 3.2}$  & $\mathbf{19.7/20}{\scriptstyle\,\pm 0.6}$   & $14/20$  & $\mathbf{20/20}$   \\
  Stand Cup Up   & $5.3/20{\scriptstyle\,\pm 1.2}$   & $\mathbf{14.7/20}{\scriptstyle\,\pm 1.2}$   & $5/20$   & $\mathbf{8/20}$    \\
  \midrule
  Average        & $7.6/20{\scriptstyle\,\pm 1.7}$  & $\mathbf{17.2/20}{\scriptstyle\,\pm 0.9}$  & $8.4/20$ & $\mathbf{15.2/20}$ \\
  \bottomrule
  \end{tabular}}
\end{minipage}\hfill
\begin{minipage}[t]{0.42\textwidth}
  \centering
  \vspace{0pt}
  \captionof{figure}{Success rates across 3-seed training in simulation. Shaded regions denote standard deviation.}
  \label{fig:training-curves}
  \includegraphics[width=\linewidth]{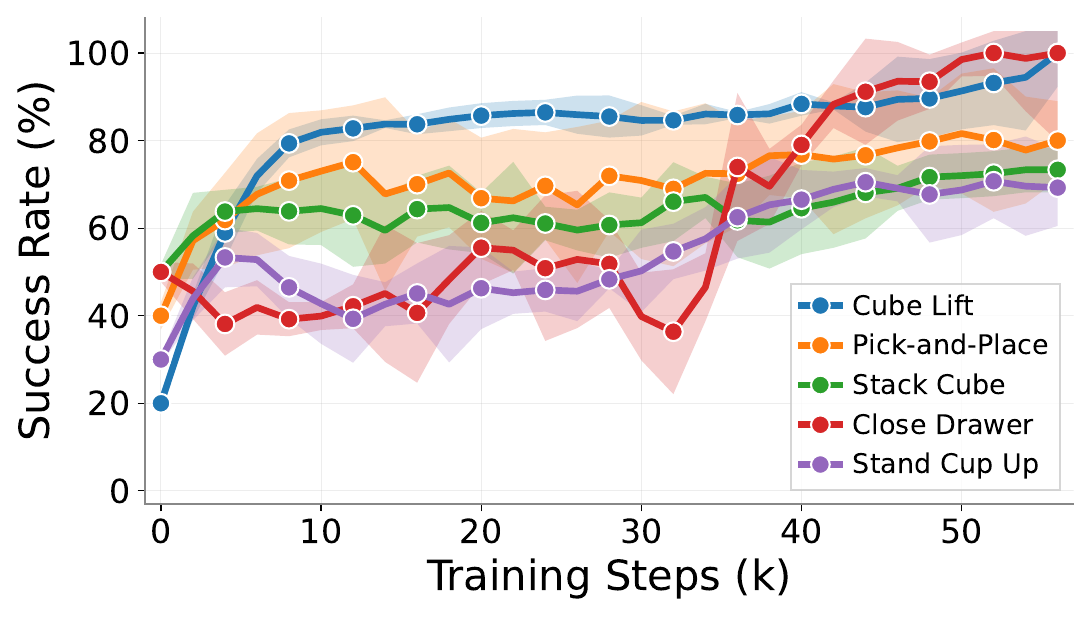}
\end{minipage}
\end{figure*}

We first examine whether a residual policy trained entirely in simulation can improve the base VLA on a real robot without any adaptation.
Table~\ref{tab:main-results} reports success rates across all five tasks. 
The residual policy improves all five tasks in simulation, with the largest gains where the base VLA struggles most.
On the real robot, the sim-trained residual transfers zero-shot to all five tasks, raising the average success rate from $42\%$ to $76\%$ without any real-world RL or fine-tuning.

\paragraph{Generalization across VLA architectures.}
To demonstrate that our residual RL framework is not specific to a single base VLA, we evaluate with $\pi_{0.5}$~\cite{pi05_2025}.
As shown in Fig.~\ref{fig:sft}(a), the residual RL consistently improves performance on the real robot, suggesting that the proposed object-centric observation interface is compatible with different VLA backbones.

\subsection{Ablations}
\label{sec:ablation}

\begin{figure*}[t]
\centering
\begin{minipage}[t]{0.65\textwidth}
  \centering
  \vspace{0pt}
  \captionof{table}{Ablation studies on real-robot performance (successes / 20 trials). \textbf{(a)} Robustness training: pose dropout and noise injection both contribute; combined training yields the strongest sim-to-real transfer. \textbf{(b)} Observation space: object-centric poses transfer best by avoiding the visual domain gap.}
  \label{tab:ablation-combined}
  \resizebox{\linewidth}{!}{%
  \begin{tabular}{lccccc}
  \toprule
   & Lift & PnP & Stack & Close & Stand \\
  \midrule
  Ours     & $\mathbf{17/20}$ & $\mathbf{16/20}$ & $\mathbf{15/20}$ & $\mathbf{20/20}$ & $\mathbf{8/20}$ \\
  \midrule
  \multicolumn{6}{l}{\textit{(a) Robustness Training}} \\
  \quad w/o noise injection      & $16/20$ & $14/20$ & $11/20$ & $\mathbf{20/20}$ & $\mathbf{8/20}$ \\
  \quad w/o pose dropout         & $13/20$ & $12/20$ & $10/20$ & $16/20$ & $7/20$ \\
  \quad w/o both                 & $12/20$ & $10/20$ & $9/20$  & $16/20$ & $5/20$ \\
  \midrule
  \multicolumn{6}{l}{\textit{(b) Observation Space}} \\
  \quad Distillation-based        & $9/20$  & $4/20$  & $8/20$  & $\mathbf{20/20}$ & $5/20$ \\
  \quad Image-based               & $8/20$  & $10/20$ & $9/20$  & $14/20$ & $6/20$ \\
  \bottomrule
  \end{tabular}}
\end{minipage}\hfill
\begin{minipage}[t]{0.3\textwidth}
  \centering
  \vspace{0pt}
  \includegraphics[width=\linewidth]{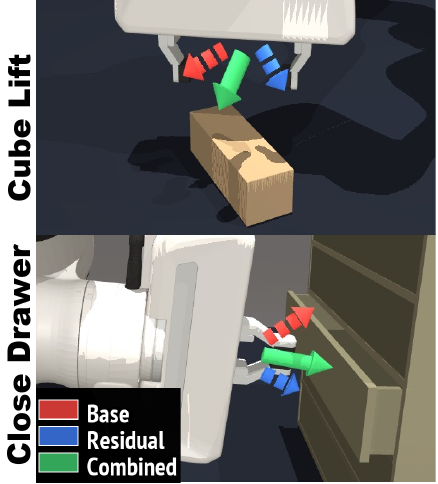}
  \captionof{figure}{The residual corrects the base action toward the goal when misaligned.
  }
  \label{fig:action-correction}
\end{minipage}
\end{figure*}

\paragraph{Robustness training.}
Table~\ref{tab:ablation-combined}(a) ablates the two robustness mechanisms from Section~\ref{sec:robustness}: pose dropout contributes most strongly
(resilience to detection failures), noise injection helps tight-tolerance tasks, and combining both yields the strongest transfer.

\paragraph{Observation space.}
\label{sec:ablation-obs}
Table~\ref{tab:ablation-combined}(b) and Fig.~\ref{fig:sft}(b) compare three observation designs. The image-based baseline suffers from the visual sim-to-real gap, and the distillation baseline, which distills a privileged-state teacher into an image-based student, loses performance during distillation. In contrast, our object-centric residual transfers best, indicating that the visual domain gap is the dominant sim-to-real barrier, which our object-centric observation sidesteps by design.

\subsection{Analysis}
\label{sec:analysis}

\begin{figure}[t]
    \centering
    \includegraphics[width=0.49\linewidth]{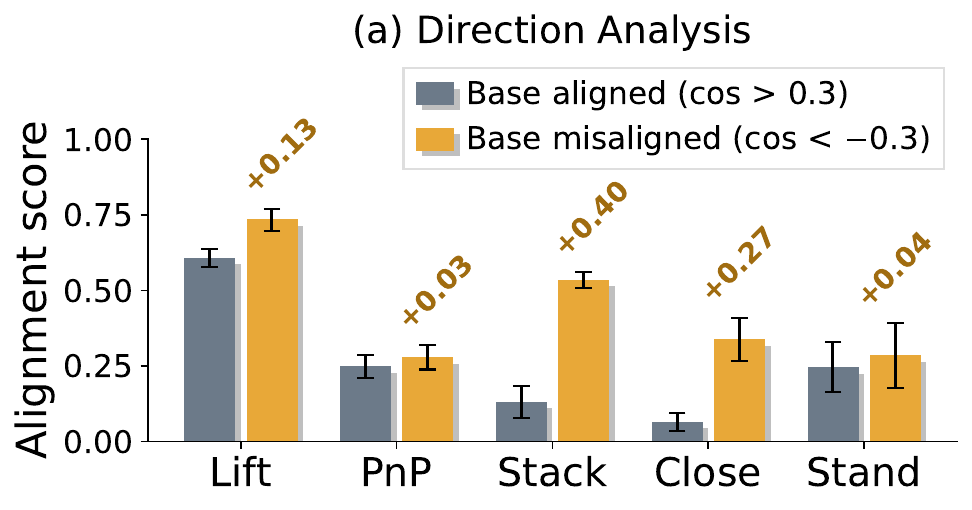}
    \hfill
    \includegraphics[width=0.49\linewidth]{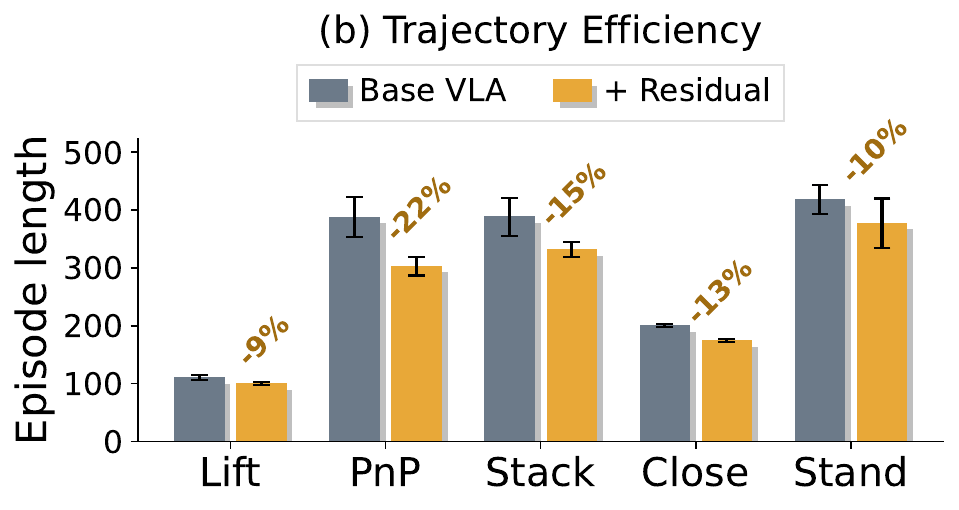}
    \caption{\textbf{(a)} Cosine similarity between the residual action and the goal direction, conditioned on base action alignment. The residual corrects more strongly when the base deviates. \textbf{(b)} Episode length comparison between base and residual-corrected policies (success episodes). The residual consistently reduces completion time by $9$--$22$\%. Error bars denote standard error of the mean across timesteps~(a) and episodes~(b).
    }
\label{fig:analysis}
\end{figure}

Having established that the residual improves performance, we now investigate when and how it intervenes.
Fig.~\ref{fig:action-correction} visualizes per-step action vectors for two tasks: the residual steers the combined action toward the goal when the base is misaligned. Across all five real-robot tasks (Fig.~\ref{fig:analysis}(a)), the residual consistently points toward the goal when the base is misaligned and contributes less when aligned, confirming selective correction; this translates to $9$--$22\%$ faster task completion (Fig.~\ref{fig:analysis}(b)).


\subsection{VLA Self-Improvement}
\label{sec:sft}
\begin{figure}[t]
\centering
\includegraphics[width=\linewidth]{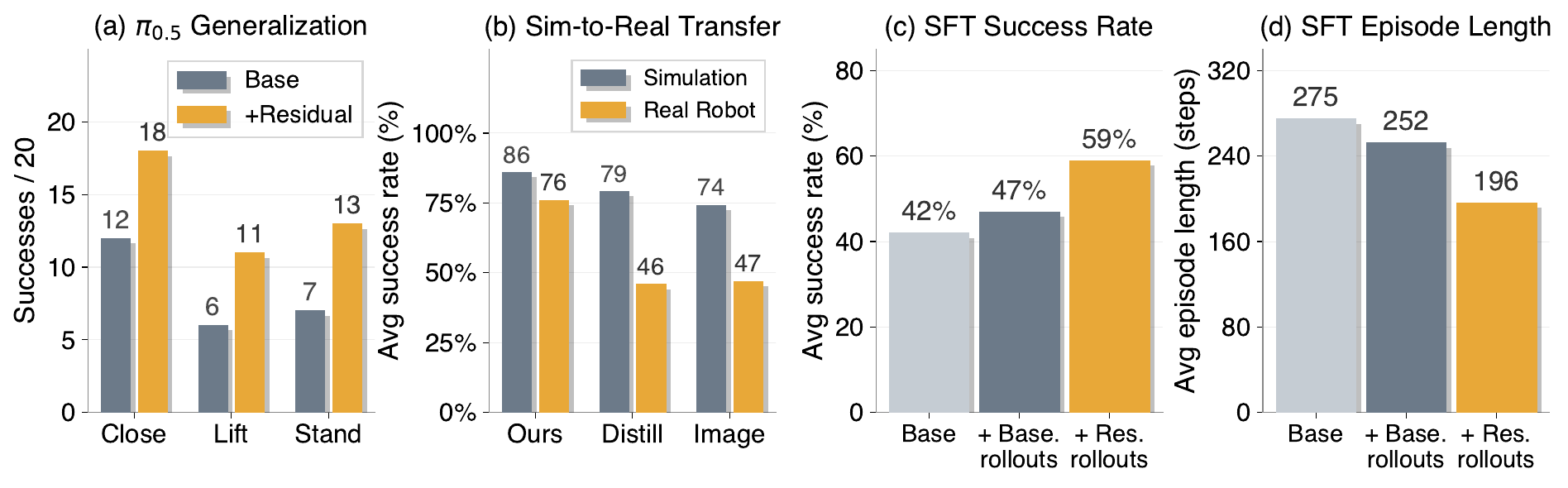}
\caption{\textbf{(a)} Performance improvement on $\pi_{0.5}$~\cite{pi05_2025}, demonstrating compatibility with different VLA backbones. \textbf{(b)} Sim-to-real transfer across observation spaces; the object-centric design transfers most effectively. \textbf{(c, d)} SFT on residual-corrected rollouts improves success rate and reduces episode length.}
\label{fig:sft}
\end{figure}

Beyond direct deployment, supervised fine-tuning (SFT) of the base VLA on residual-corrected rollouts raises real-robot success rate and reduces episode length compared to SFT on plain base rollouts (Fig.~\ref{fig:sft}(c,d)). These results suggest that residual-corrected trajectories provide higher-quality supervision for retraining the base VLA, enabling a self-improvement loop without additional teleoperation.

\section{Conclusion}
\label{sec:conclusion}
We presented object-centric residual RL, a method for enhancing Vision-Language-Action models through sim-trained residual policies that transfer to the real robot zero-shot.
The key insight is that an object-centric observation space, constructed from 6-DoF object poses, proprioception, and the base VLA action, is recoverable in both simulation and reality without visual rendering, enabling strong zero-shot sim-to-real transfer of the residual policy. Combined with robustness training via noise injection and pose dropout, our residual policies improve a GR00T-N1.5-based VLA across five tasks in both simulation and on a real FR3 robot, raising the real-robot average success rate from $42\%$ to $76\%$ without any real-world reinforcement learning or residual-policy fine-tuning. Beyond direct deployment, the residual-corrected policy generates improved real-robot rollouts that can be aggregated across tasks to retrain a single multi-task VLA, enabling a self-improvement loop that requires no additional teleoperation.
We believe this paradigm, which combines the generalization of VLAs with the precise corrective capability of RL through a carefully chosen observation interface, offers a practical path toward scalable and autonomous robot improvement.

\section{Limitations and Future Work}
\label{sec:limitations}
Our method relies on real-time 6-DoF pose tracking (FoundationPose + SAM2), which can fail under full occlusion or heavy clutter; memory-based pose estimation could mitigate this limitation. Task-relevant objects must also be specified manually; scaling to open-world settings would require automatic identification, e.g., from VLA attention maps.
The pose-based observation bridges the visual domain gap but not the dynamics gap: contact friction and gripper compliance differences between sim and real may cause suboptimal corrections in contact-rich tasks.
As a residual architecture, the policy can correct mild deviations but cannot recover from states far outside the base VLA's training distribution.
Finally, tasks requiring sub-millimeter precision or involving very small objects may exceed the accuracy of current pose estimation; extending to such scenarios with higher-resolution sensing or tactile feedback is left to future work.


\clearpage

\clearpage
\bibliography{reference}


\ifincludeappendix
\clearpage
\begin{center}
{\Large \bfseries Appendix for:\\[0.3em] Object-Centric Residual RL for Zero-Shot Sim-to-Real VLA Enhancement\par}
\end{center}
\vspace{0.5em}
\appendix
\section{Appendix}
\label{sec:appendix}
\subsection{Reward Design}
\label{sec:app-reward}
All tasks use dense, shaped rewards clipped to $[0, 1]$.
Each reward is decomposed into staged sub-rewards that are applied progressively as the task advances.
Table~\ref{tab:rewards} summarizes the reward structure per task.

\begin{table}[h]
\centering
\caption{Reward stages per task. Each stage provides a continuous signal based on distance, orientation, or contact metrics.}
\label{tab:rewards}
\small
\begin{tabular}{ll}
\toprule
Task & Reward stages \\
\midrule
Cube Lift & Reach $\rightarrow$ Grasp $\rightarrow$ Lift \\
Pick-and-Place & Reach $\rightarrow$ Grasp $\rightarrow$ Carry $\rightarrow$ Place \\
Stack Cube & Reach $\rightarrow$ Grasp $\rightarrow$ Align $\rightarrow$ Stack \\
Close Drawer & Push (closing progress) $\rightarrow$ Close (fully closed) \\
Stand Cup Up & Reach $\rightarrow$ Grasp $\rightarrow$ Upright \\
\bottomrule
\end{tabular}
\end{table}

\subsection{Simulation Environment Construction}
\label{sec:app-sim-construction}

All tasks are built in MuJoCo~\cite{todorov2012mujoco}.
Object dimensions and workspace layout are measured from the real setup; other scene parameters (table height, camera pose, etc.) do not require exact matching since the sim and real VLAs are trained separately and the residual policy does not observe images.
Each object is modeled using a simple geometric primitive from the measured principal dimensions.
For Cube Lift, Pick-and-Place, and Stack Cube, the cube is modeled as a simple box with measured side length; Pick-and-Place additionally includes a bowl, and Stack Cube places two cubes.
For Close Drawer, the cabinet and drawer are modeled with a sliding joint whose range matches the real drawer travel.
For Stand Cup Up, the cup is modeled as a truncated cone (approximated by stacked cylinder slices) placed on its side as the initial pose.
Table~\ref{tab:sim-objects} summarizes the object specifications used in simulation.
During RL training, the initial position and orientation of task-relevant objects are randomized within the workspace to expose the residual policy to diverse configurations.

\begin{table}[h]
\centering
\caption{Simulation object specifications. All dimensions are measured from the real objects; masses correspond to the values used in simulation. All cuboid blocks (Cube Lift, Pick-and-Place, Stack Cube) and the Close Drawer cabinet/drawers are modeled with a wood material; the Pick-and-Place bowl is a thin paper bowl; the Stand Cup Up cup is a rigid plastic-like shell.
}
\label{tab:sim-objects}
\small
\resizebox{\textwidth}{!}{%
\begin{tabular}{llllll}
\toprule
Task & Object & Geometry & Dimensions & Mass & Color \\
\midrule
Cube Lift & Cuboid & box & $12 \times 4 \times 4$\,cm & $75$\,g & wood brown \\
\midrule
\multirow{2}{*}{Pick-and-Place} & Cuboid (grasped) & box & $8 \times 4 \times 4$\,cm & $50$\,g & red \\
 & Bowl (target) & cylinder & radius $9.5$\,cm, height $1$\,cm & $10$\,g & white \\
\midrule
\multirow{2}{*}{Stack Cube} & Cube (grasped) & box & $4 \times 4 \times 4$\,cm & $25$\,g & white \\
 & Cube (target) & box & $4 \times 4 \times 4$\,cm & $25$\,g & green \\
\midrule
\multirow{2}{*}{Close Drawer} & Cabinet & box composite & $27 \times 35 \times 28.2$\,cm & $4.9$\,kg & wood brown \\
 & Drawer ($\times 5$) & slide joint & travel $13$\,cm & $0.87$\,kg & wood brown \\
\midrule
Stand Cup Up & Cup & truncated cone & top $\varnothing\,7.5$\,cm, bottom $\varnothing\,5$\,cm, height $10$\,cm & $150$\,g & red \\
\bottomrule
\end{tabular}%
}
\end{table}

\subsection{Realistic Simulation Rendering}
\label{sec:app-realistic-sim}

While our main method uses object-centric observations, we additionally set up a visually realistic MuJoCo rendering environment to enable image-based RL training and policy distillation as auxiliary baselines in simulation.
Specifically, we use high-quality textured meshes for the robot, objects, and tabletop; physically-based lighting with area lights matching the lab illumination; and domain randomization~\cite{tobin2017domain} over backgrounds, lighting intensity, and camera pose.
Fig.~\ref{fig:realistic-sim} shows side-by-side comparisons of the real-world camera view (left) and the corresponding simulation rendering (right) across all five tasks.

\begin{figure}[h]
    \centering
    \begin{subfigure}[t]{0.48\linewidth}
        \includegraphics[width=\linewidth]{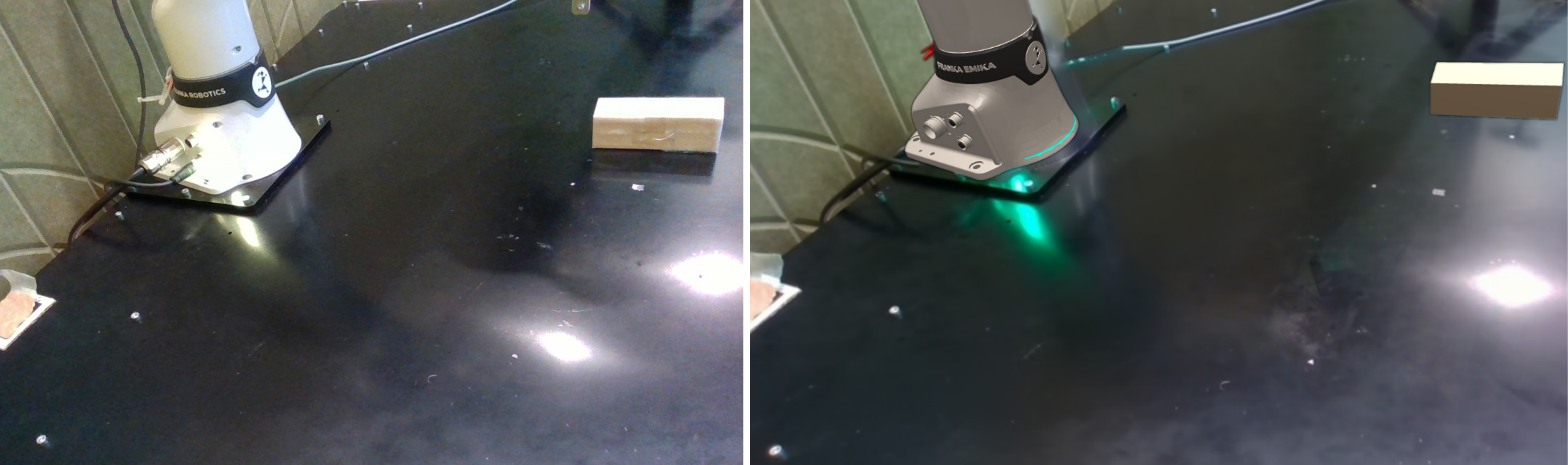}
        \caption{Cube Lift}
    \end{subfigure}\hfill
    \begin{subfigure}[t]{0.48\linewidth}
        \includegraphics[width=\linewidth]{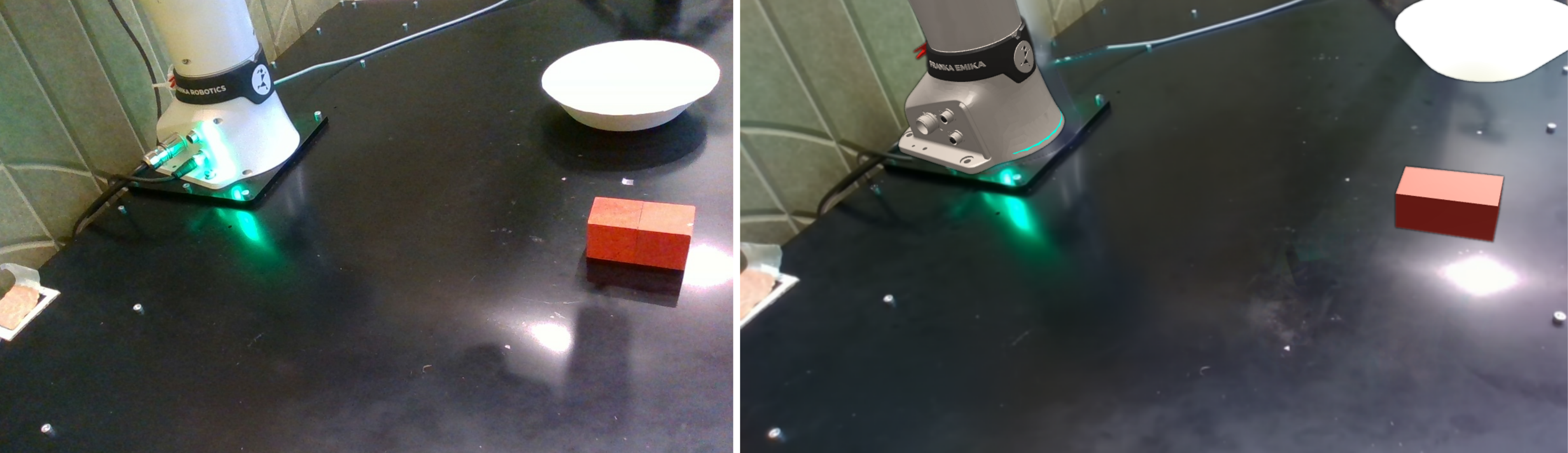}
        \caption{Pick-and-Place}
    \end{subfigure}\\[4pt]
    \begin{subfigure}[t]{0.48\linewidth}
        \includegraphics[width=\linewidth]{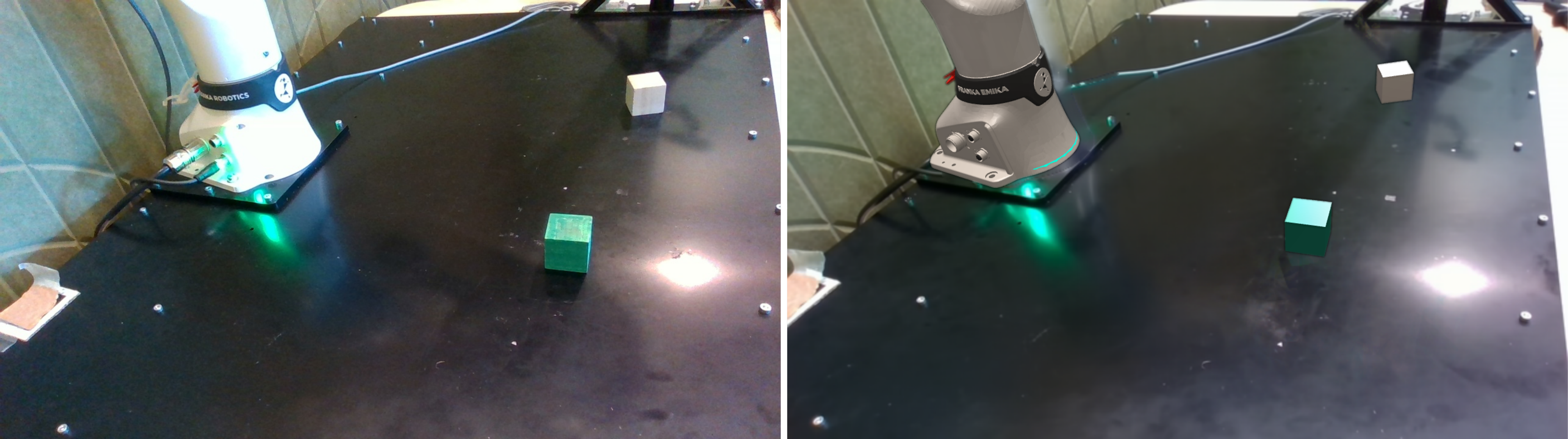}
        \caption{Stack Cube}
    \end{subfigure}\hfill
    \begin{subfigure}[t]{0.48\linewidth}
        \includegraphics[width=\linewidth]{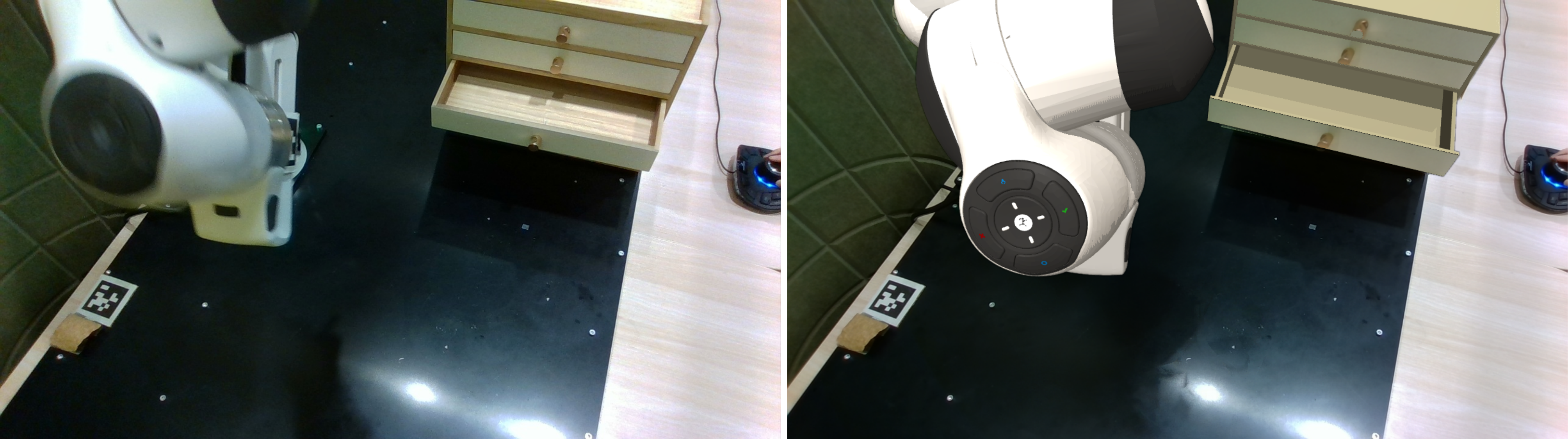}
        \caption{Close Drawer}
    \end{subfigure}\\[4pt]
    \begin{subfigure}[t]{0.48\linewidth}
        \includegraphics[width=\linewidth]{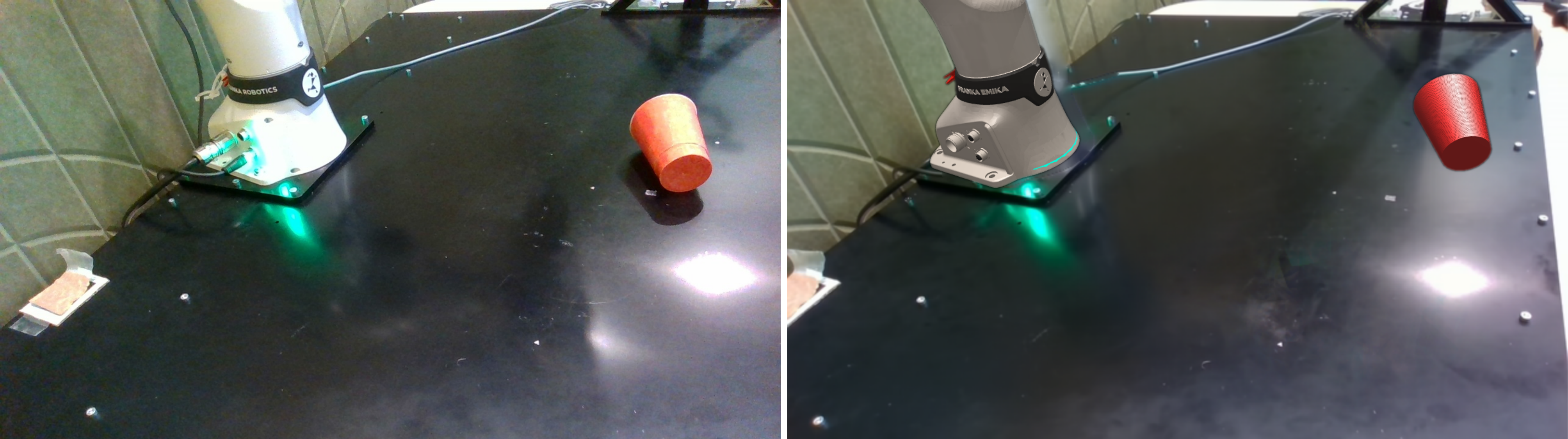}
        \caption{Stand Cup Up}
    \end{subfigure}
    \caption{Realistic simulation rendering (right in each pair) vs.\ real-world camera view (left). The rendering is set up to support image-based RL and policy distillation as auxiliary baselines in simulation.}
    \label{fig:realistic-sim}
\end{figure}

\subsection{Algorithm Pseudocode}
\label{sec:app-algorithms}

Algorithms~\ref{alg:training} and~\ref{alg:inference} provide complete pseudocode for the two phases of our framework. Algorithm~\ref{alg:training} details the residual RL training loop in simulation, including VLA action chunking, pose noise augmentation, and TD3~\cite{fujimoto2018td3} updates. Algorithm~\ref{alg:inference} describes the zero-shot real-world deployment procedure, where a confidence-gated pose dropout replaces the stochastic dropout used during training.
The pose estimator combines FoundationPose~\cite{wen2024foundationpose} for 6-DoF tracking with SAM2~\cite{ravi2024sam2} for instance segmentation.

\begin{algorithm}[h]
\caption{Object-Centric Residual RL Training}
\label{alg:training}
\begin{algorithmic}[1]
\Require Frozen $\pi_\mathrm{VLA}^{\mathrm{sim}}$, language instruction $l$, episode length $T$, chunk length $H$, noise parameters $\sigma_p^{\max}, \sigma_q^{\max}, \rho_{\text{drop}}$, exploration noise std $\tilde{\sigma}$, clip bound $c$
\State Initialize actor $\pi_{\text{res}}^{\text{sim}}$, critics $Q_1, Q_2$, target networks, replay buffer $\mathcal{B}$
\For{each episode}
    \State $s_0 \leftarrow \text{env.reset}()$
    \For{$t = 0, 1, \ldots, T-1$}
        \If{$t \bmod H = 0$}
            \State $A \leftarrow \pi_\mathrm{VLA}^{\mathrm{sim}}(o_t^{\text{img}}, s_t^{\text{prop}}, l)$ \Comment{Query frozen VLA, $H$-length chunk}
        \EndIf
        \State $a_t^{\text{base}} \leftarrow A[t \bmod H]$
        \State $\tilde{x}_{\text{obj}} \leftarrow \Call{PoseAugment}{p_{\text{obj}}, q_{\text{obj}}, \sigma_p^{\max}, \sigma_q^{\max}, \rho_{\text{drop}}}$
        \State $s_t \leftarrow [\,\tilde{x}_{\text{obj}},\; s_t^{\text{prop}},\; a_t^{\text{base}}\,]$
        \State $\delta_t \leftarrow \pi_{\text{res}}^{\text{sim}}(s_t) + \epsilon$, \quad $\epsilon \sim \text{clip}(\mathcal{N}(0, \tilde{\sigma}), -c, c)$ \Comment{Noise on residual}
        \State $a_t \leftarrow a_t^{\text{base}} \oplus \delta_t$
        \State Execute $a_t$ in env, observe $r_t, s_{t+1}$
        \State $\mathcal{B} \leftarrow \mathcal{B} \cup \{(s_t, a_t, r_t, s_{t+1})\}$ \Comment{Store combined action}
        \State Sample mini-batch from $\mathcal{B}$; update $Q_1, Q_2$ and $\pi_{\text{res}}^{\text{sim}}$ (TD3)
    \EndFor
\EndFor
\State \Return $\pi_{\text{res}}^{\text{sim}}$
\end{algorithmic}
\end{algorithm}

\begin{algorithm}[h]
\caption{Zero-Shot Real-World Deployment}
\label{alg:inference}
\begin{algorithmic}[1]
\Require Frozen $\pi_\mathrm{VLA}^{\mathrm{real}}$, frozen $\pi_{\text{res}}^{\text{sim}}$, language instruction $l$, chunk length $H$, pose-confidence threshold $\tau_c$
\For{$t = 0, 1, \ldots$ until task completion}
    \State $o_t^{\text{img}} \leftarrow \text{CaptureRGB}()$
    \If{$t \bmod H = 0$}
        \State $A \leftarrow \pi_\mathrm{VLA}^{\mathrm{real}}(o_t^{\text{img}}, s_t^{\text{prop}}, l)$ \Comment{Query frozen VLA, $H$-length chunk}
    \EndIf
    \State $a_t^{\text{base}} \leftarrow A[t \bmod H]$
    \State $\tilde{x}_{\text{obj}}, c_t \leftarrow \text{FoundationPose}(\text{SAM2}(o_t^{\text{img}}))$ \Comment{6-DoF pose + confidence}
    \If{$c_t < \tau_c$} $\tilde{x}_{\text{obj}} \leftarrow \mathbf{0}$ \Comment{Low confidence $\to$ dropout}
    \EndIf
    \State $s_t \leftarrow [\,\tilde{x}_{\text{obj}},\; s_t^{\text{prop}},\; a_t^{\text{base}}\,]$
    \State $a_t \leftarrow a_t^{\text{base}} \oplus \pi_{\text{res}}^{\text{sim}}(s_t)$
    \State Execute $a_t$ on robot
\EndFor
\end{algorithmic}
\end{algorithm}

\subsection{Training Hyperparameters}
\label{sec:app-hparams}

Table~\ref{tab:hparams} lists the task-specific hyperparameters used for residual TD3~\cite{fujimoto2018td3} training.
All tasks share the same network architecture (2-layer MLP with $512$-unit actor and $1024$-unit critic hidden layers) and optimizer (Adam). Per-task differences in learning rate, L2 regularization, discount factor, episode length, offline sampling fraction, and critic warmup accommodate the varying task dynamics.

\begin{table}[h]
\centering
\caption{Task-specific training hyperparameters.}
\label{tab:hparams}
\small
\begin{tabular}{lccccc}
\toprule
& Lift & PnP & Stack & Close & Stand \\
\midrule
Actor hidden dim & $512$ & $512$ & $512$ & $512$ & $512$ \\
Critic hidden dim & $1024$ & $1024$ & $1024$ & $1024$ & $1024$ \\
Actor LR & $1\mathrm{e}{\text{-}5}$ & $1\mathrm{e}{\text{-}5}$ & $3\mathrm{e}{\text{-}4}$ & $1\mathrm{e}{\text{-}5}$ & $1\mathrm{e}{\text{-}5}$ \\
Critic LR & $1\mathrm{e}{\text{-}4}$ & $1\mathrm{e}{\text{-}4}$ & $3\mathrm{e}{\text{-}4}$ & $1\mathrm{e}{\text{-}4}$ & $1\mathrm{e}{\text{-}4}$ \\
L2 reg. & $0.0$ & $0.1$ & $1.0$ & $0.01$ & $1.0$ \\
$\gamma$ & $0.99$ & $0.99$ & $0.95$ & $0.99$ & $0.99$ \\
Batch size & $256$ & $256$ & $256$ & $256$ & $256$ \\
Offline fraction & $0.5$ & $0.5$ & $0.75$ & $0.3$ & $0.5$ \\
Max episode steps & $300$ & $300$ & $500$ & $500$ & $300$ \\
Critic warmup steps & $1\mathrm{k}$ & $1\mathrm{k}$ & $2\mathrm{k}$ & $1\mathrm{k}$ & $1\mathrm{k}$ \\
$n$-step returns & $3$ & $3$ & $3$ & $3$ & $3$ \\
Target $\tau$ & $0.005$ & $0.005$ & $0.005$ & $0.005$ & $0.005$ \\
Exploration noise $\tilde{\sigma}$ & $0.05$ & $0.05$ & $0.05$ & $0.05$ & $0.05$ \\
\bottomrule
\end{tabular}
\end{table}

\subsection{Pose Noise and Dropout Parameters}
\label{sec:app-pose-noise}

We use $\sigma_p^{\max} = 0.005$ ($5$\,mm) for position noise and $\sigma_q^{\max} = 0.1$\,rad ($\approx 5.7^\circ$) for orientation noise.
Each component of $\epsilon_p$ is independently sampled from $\mathcal{U}(-\tilde{\sigma}_p, \tilde{\sigma}_p)$ with $\tilde{\sigma}_p \sim \mathcal{U}(0, \sigma_p^{\max})$ resampled per timestep, and $\epsilon_q$ is a small random rotation whose magnitude is similarly drawn via $\tilde{\sigma}_q \sim \mathcal{U}(0, \sigma_q^{\max})$.
These ranges match the typical depth-camera-based pose estimation error commonly reported for Intel RealSense D435 (${\sim}$2.5--5\,mm at 1\,m distance) and the orientation jitter of FoundationPose~\cite{wen2024foundationpose} under nominal tracking conditions.
During training, pose dropout is applied with probability $\rho_{\text{drop}} = 0.1$.

\paragraph{Additional fixed constants.} We use action chunk length $H = 16$ (the GR00T-N1.5~\cite{bjorck2025gr00t} default), exploration noise clip bound $c = 0.5$, and pose-confidence threshold $\tau_c = 0.5$ at deployment.

\subsection{Sim-to-Real Behavioral Consistency}
\label{sec:app-sim-real-behavior}

A key validation of our zero-shot transfer framework is that base VLA failure modes observed on the real robot are reproduced in simulation: because $\pi_\mathrm{VLA}^{\mathrm{sim}}$ and $\pi_\mathrm{VLA}^{\mathrm{real}}$ are trained on the same teleoperation data, they exhibit the same characteristic failures (e.g., hovering above the cube, stopping short of the target).
Training the residual against these shared failures in simulation thus directly addresses the failures encountered on the real robot, consistent with the premise of residual policy learning~\cite{silver2018residual, johannink2019residual}.
Table~\ref{tab:behavior-summary} summarizes the shared failure modes and the corresponding residual corrections (trained in sim, deployed zero-shot to real).

\begin{table}[h]
\centering
\caption{Base VLA failure modes shared between simulation and the real robot, and the residual corrections that resolve them (learned in simulation, deployed zero-shot to real).}
\label{tab:behavior-summary}
\small
\begin{tabular}{lll}
\toprule
Task & Base VLA Problem & Residual Fix \\
\midrule
\multirow{3}{*}{\shortstack[l]{Lift, PnP,\\Stack}} & Hovers above cube, misses grasp & Pushes end-effector down to the cube \\
& Stops short of target position/angle & Moves end-effector closer to the target \\
& Gripper collides with cube and stalls & Adjusts approach angle \\
\midrule
Close Drawer & Pushes at wrong angle & Redirects along drawer axis \\
\midrule
Stand Cup Up & End-effector can't reach correct grasp pose & Guides end-effector to precise grasp position \\
\bottomrule
\end{tabular}
\end{table}

These observations confirm the central premise of our framework: when paired sim/real VLAs share the same failure distribution, a residual trained to correct the sim failures resolves the same failures on the real robot zero-shot, without observing images at any point.
Fig.~\ref{fig:sim-real-behavior} shows representative keyframe comparisons across all five tasks.

\begin{figure}[h]
    \centering
    \includegraphics[width=\linewidth]{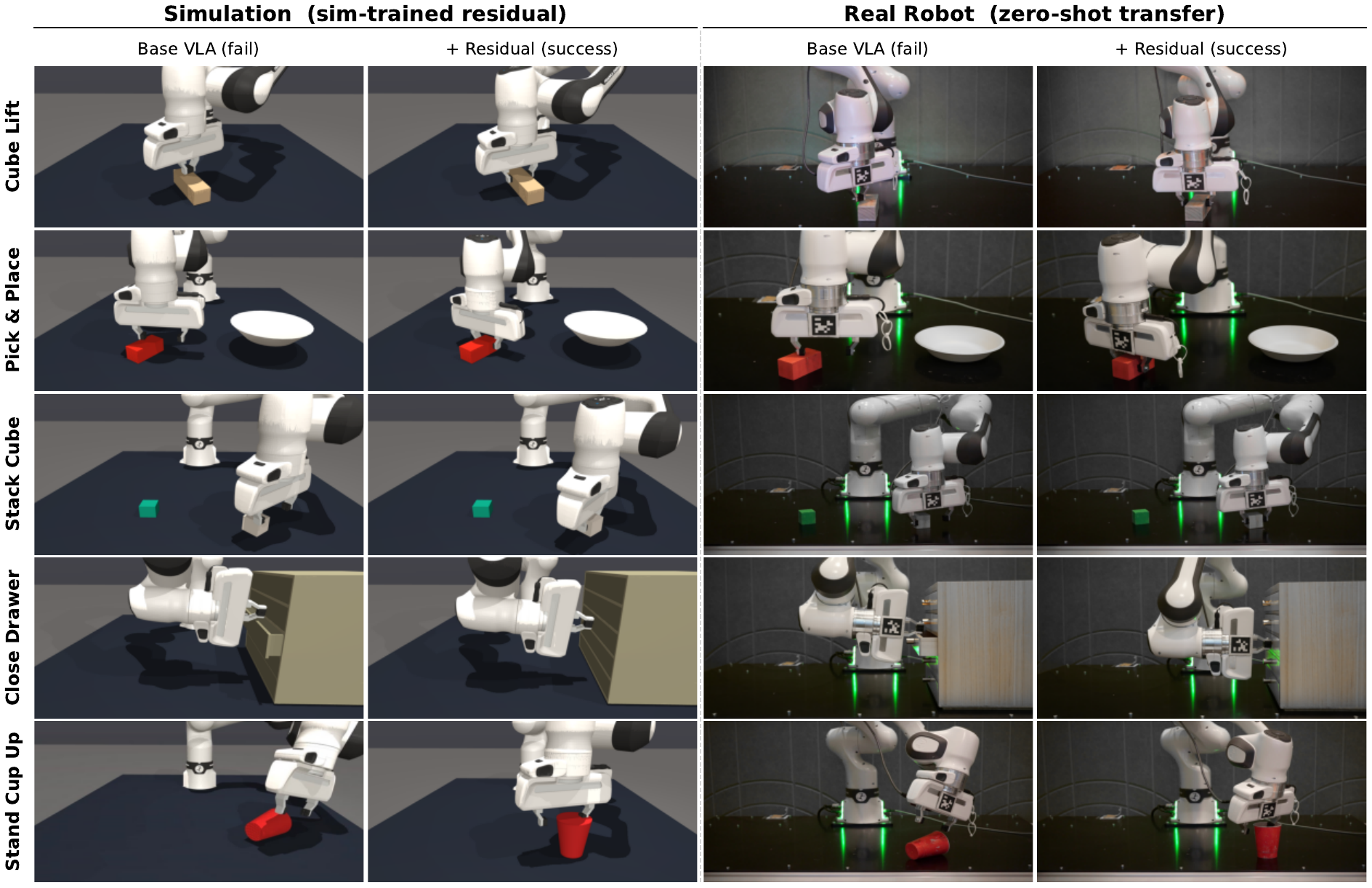}
    \caption{\textbf{Sim-to-real behavioral transfer of the object-centric residual policy.} All five tasks are shown; left two columns are simulation, right two are real-robot deployment. The residual, trained \emph{only} in simulation, learns task-specific corrections to base VLA failure modes: downward correction during \textit{Cube Lift} approach, lateral alignment for \textit{Pick-and-Place}, accurate grasp positioning for \textit{Stack Cube}, corrective pushing motion to fully close the drawer in \textit{Close Drawer}, and re-orientation to stand the cup upright in \textit{Stand Cup Up}. Because the residual observes object pose, a representation invariant across domains, the same corrections transfer zero-shot to the real robot.}
    \label{fig:sim-real-behavior}
\end{figure}

\subsection{Strong Base VLA + Residual RL}
\label{sec:app-strong-base}

A natural question is whether residual RL still helps when the base VLA is already strong.
We investigate this by comparing two tiers of base VLA performance on the Pick-and-Place task (Table~\ref{tab:strong-base}).

\begin{table}[h]
\centering
\caption{Tier 1 vs.\ Tier 2: Residual RL on weak and strong base VLAs (Pick-and-Place, 20 trials).}
\label{tab:strong-base}
\small
\begin{tabular}{llcc}
\toprule
 & Method & Sim SR & Real SR \\
\midrule
\multirow{2}{*}{Tier 1 (Weak, GR00T-N1.5)}  & Base VLA       & $7/20$ & $9/20$ \\
                                      & + Residual RL  & $\mathbf{17/20}$ & $\mathbf{16/20}$ \\
\midrule
\multirow{2}{*}{Tier 2 (Strong, $\pi_{0.5}$)} & Base VLA       & $18/20$ & $\mathbf{17/20}$ \\
                                      & + Residual RL  & $\mathbf{19/20}$ & $\mathbf{17/20}$ \\
\bottomrule
\end{tabular}
\end{table}

With a weak base (Tier 1), residual RL roughly doubles the success rate in both simulation and the real world.
With a strong base (Tier 2, $\pi_{0.5}$~\cite{pi05_2025}, 17/20 real), the residual maintains the same real-world success rate without degradation, while slightly improving simulation performance.
This is the expected behavior: when the base already acts correctly, the residual learns to stay near zero and does not introduce unnecessary corrections.

However, residual RL remains useful for two reasons.
First, VLA success rates are highly dependent on the deployment environment---a model that appears strong in one setting can become a weak base in another, and residual RL provides a safety net in such cases.
Second, when the base is weak, it is often unclear what additional demonstration data would make it stronger.
Residual RL sidesteps this problem: the policy explores diverse situations in simulation and autonomously discovers failure modes that human teleoperators may not anticipate when curating fine-tuning data.
Moreover, unlike real-robot data collection where performance cannot be easily evaluated during the process, sim-based residual RL allows continuous evaluation and monitoring of improvement throughout training.
Finally, our sim-trained residual is complementary to recent VLAs that learn from execution experience~\cite{pi06_2025}: it can be applied on top of any frozen base policy without modifying the base's training pipeline.

\subsection{Emergent Behaviors from Residual RL}
\label{sec:app-emergent}

Beyond closing the simulation-to-real gap, residual RL also induces qualitatively new behaviors that are absent from the demonstration data used to train the base VLA~\cite{bjorck2025gr00t}. Fig.~\ref{fig:emergent-behaviors} shows four such examples observed during real-robot deployment.

\textbf{Cube Lift (\textit{Cube Pre-Rotation}).} Before grasping, the residual nudges the cube into a graspable orientation, a strategy not present in the demonstrations.

\textbf{Pick-and-Place (\textit{Cube Pre-Rotation}).} The residual exhibits the same pre-rotation strategy before grasping the cube to place it in the bowl.

\textbf{Stack Cube (\textit{Corrective Push to Grasp}).} When the base policy's grasp is misaligned, the residual drives the gripper toward the cube to reach a full close.

\textbf{Close Drawer (\textit{Sustained Contact Push}).} The residual maintains downward contact through the late phase, avoiding the base policy's premature lift that would otherwise lose contact with the drawer.

These behaviors emerge purely from RL exploration in simulation: the policy autonomously discovers corrective strategies that human teleoperators may not anticipate when curating demonstration data, supporting the second motivation discussed in Sec.~\ref{sec:app-strong-base}.

\begin{figure}[h]
    \centering
    \includegraphics[width=\linewidth]{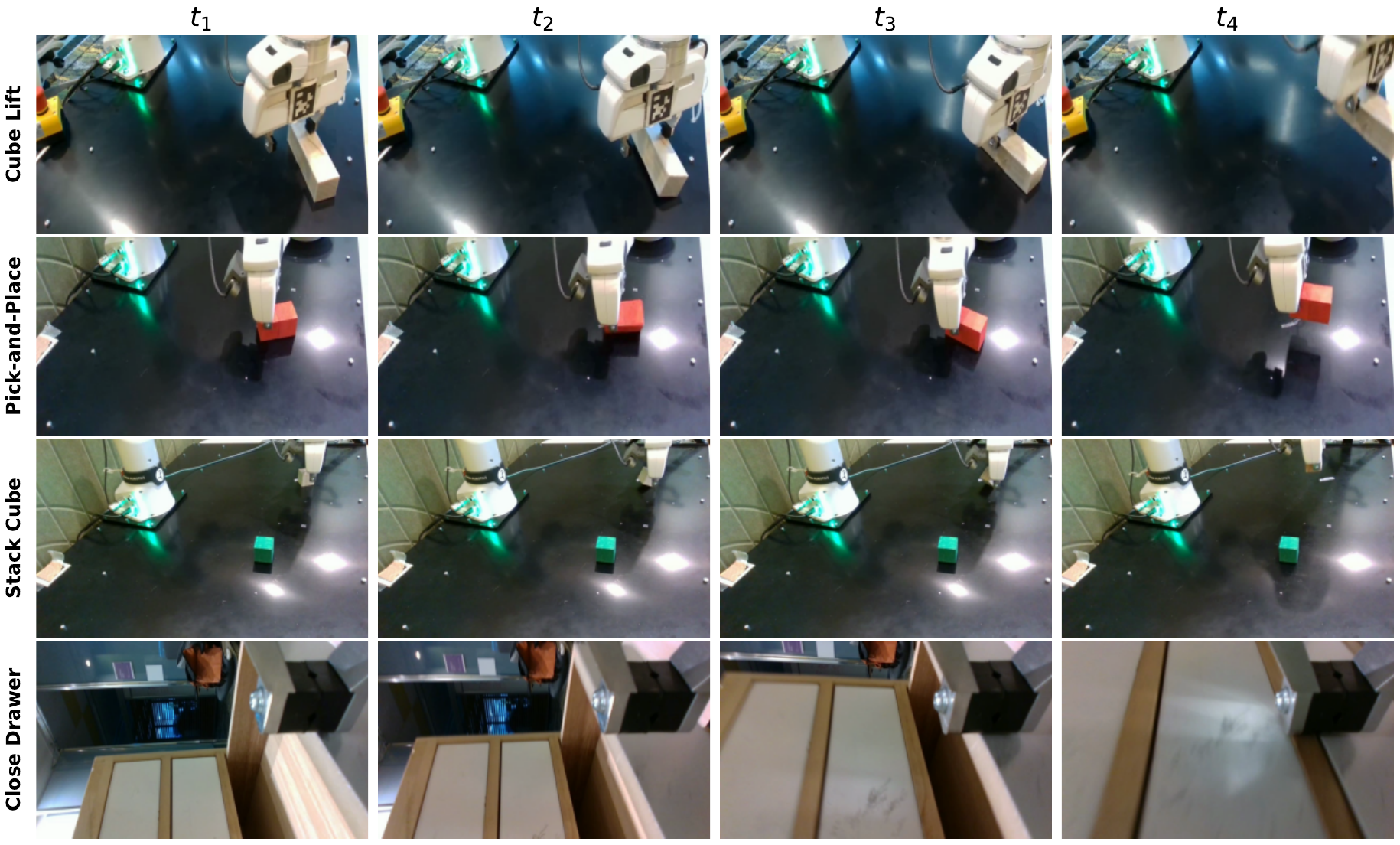}
    \caption{\textbf{Emergent behaviors from residual RL.} Each row shows four sequential keyframes (left to right in time) from a successful real-robot rollout with the residual policy. The residual discovers task-specific strategies that are absent from the demonstrations used to train the base VLA: pre-rotating the cube before grasp (\textit{Cube Lift} and \textit{Pick-and-Place}), corrective push toward the cube when the grasp is misaligned (\textit{Stack Cube}), and sustained downward contact through the late phase of drawer closing (\textit{Close Drawer}).}
    \label{fig:emergent-behaviors}
\end{figure}

\subsection{Object Tracking Visualization}
\label{sec:app-object-tracking}

We visualize the FoundationPose~\cite{wen2024foundationpose} object tracking results during real-robot deployment.
Fig.~\ref{fig:pose-tracking} shows representative frames with the estimated 6-DoF pose visualized as a projected mesh outline of each object, for all five manipulation tasks.
For Close Drawer, the policy observes the slide displacement; we visualize the closing progress using a depth-based quad detector that identifies the active drawer and estimates its slide position from the median depth within each drawer face region.

\begin{figure}[h]
    \centering
    \includegraphics[width=\linewidth]{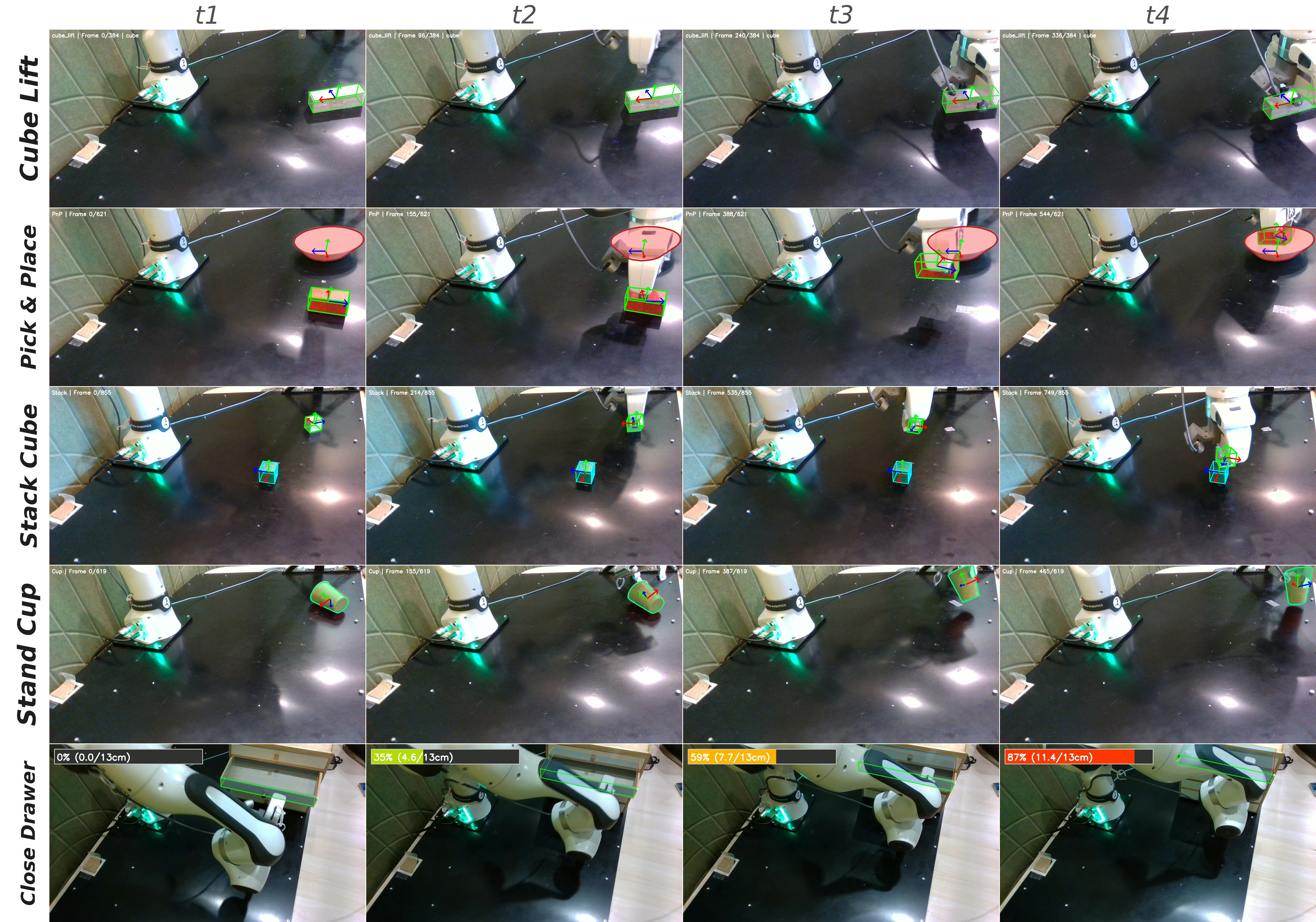}
    \caption{FoundationPose~\cite{wen2024foundationpose} pose tracking overlaid on real-robot RGB frames. Each row shows a different task (Cube Lift, Pick-and-Place, Stack Cube, Stand Cup Up, and Close Drawer) at evenly spaced timesteps during a successful episode. For each object, the estimated 6-DoF pose is visualized by projecting the object's mesh outline into the camera frame, with body-frame axes (red, green, blue) drawn at the object origin. For Close Drawer, the overlay additionally shows the closing-progress percentage on the active drawer face.}
    \label{fig:pose-tracking}
\end{figure}

\clearpage
\subsection{Failure Case Analysis}
\label{sec:app-failure-modes}

Fig.~\ref{fig:failure-cases} shows three representative failure modes observed during real-robot evaluation, with the FoundationPose~\cite{wen2024foundationpose} 6-DoF estimate overlaid on each frame.

\begin{itemize}[nosep]
    \item \textbf{Pose-estimation error} (Rows 1--2). FoundationPose returns an offset or drifted pose, causing the residual to correct in the wrong direction.
    \item \textbf{Occlusion} (Row 3). The gripper occludes the cube as it closes, and the tracker loses the object.
    \item \textbf{Wrong-object detection} (Row 4). Segmentation locks onto a specular reflection instead of the target, producing a spurious pose estimate.
\end{itemize}

All three modes stem from perception failures that pass the confidence gate. Incorporating complementary verification---such as VLM-based semantic validation or multi-hypothesis pose estimation with diverse initializations---to detect plausible-but-incorrect estimates, together with fallback strategies that gracefully degrade to base VLA behavior, is a promising direction for addressing these cases.

\begin{figure}[h]
    \centering
    \includegraphics[width=\linewidth]{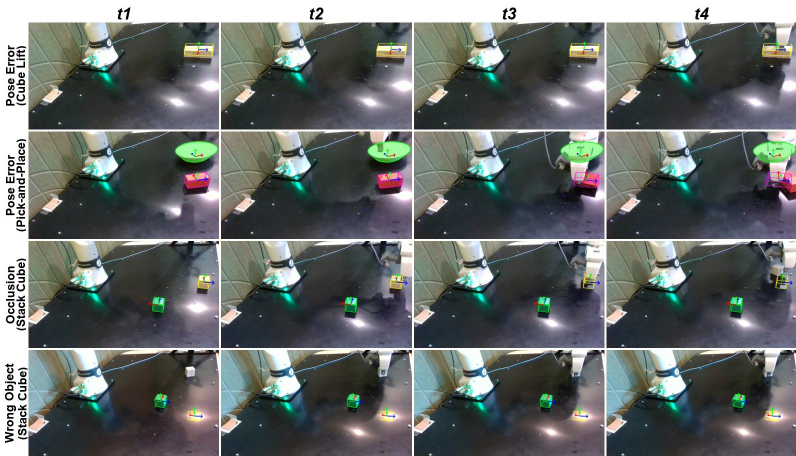}
    \caption{Representative failure modes on the real robot. Each row shows four uniformly sampled timesteps with the FoundationPose estimate overlaid. \textbf{Rows 1--2 (Pose Error)}: tracker offset on Cube Lift (Row 1) and pose drift during Pick-and-Place (Row 2). \textbf{Row 3 (Occlusion)}: gripper occludes the cube during Stack Cube grasp, causing the tracker to lose the object. \textbf{Row 4 (Wrong Object)}: segmentation locks onto a table reflection in Stack Cube.}
    \label{fig:failure-cases}
\end{figure}

\end{document}